\documentclass{article}

\usepackage{amsmath,amsfonts,amsthm,tikz,times,natbib,microtype,array,paralist,subcaption,hyperref}
\usepackage[accepted]{arxiv}

\newcommand\given{{\,|\,}}
\newcommand\BB[1]{{\mathbb{#1}}}

\newcommand{\independent}{\perp}
\renewcommand{\dot}[2]{\langle #1, #2 \rangle}
\newcommand\percent{\%}

\usepackage{mdframed}

\newtheorem{principle}{Principle}

\usetikzlibrary{arrows,calc,positioning,decorations.pathreplacing} 
\tikzset{
  >=stealth',
  vertex/.style={circle, draw=black, thick, minimum height=2em, inner sep=0pt, text centered},
  edge/.style={->, thick, shorten <=2pt, shorten >=2pt}
}

\icmltitlerunning{Causal Discovery Using Proxy Variables}

\begin{document} 

\twocolumn[
  \icmltitle{Causal Discovery Using Proxy Variables}
  \begin{icmlauthorlist}
    \icmlauthor{Mateo Rojas-Carulla}{fb,cam,mpi}
    \icmlauthor{Marco Baroni}{fb}
    \icmlauthor{David Lopez-Paz}{fb}
  \end{icmlauthorlist}
  
  \icmlaffiliation{fb}{Facebook AI Research, Paris, France}
  \icmlaffiliation{cam}{University of Cambridge, Cambridge, UK}
  \icmlaffiliation{mpi}{MPI for Intelligent Systems, T\"ubingen, Germany}

  \icmlcorrespondingauthor{Mateo Rojas-Carulla}{mrojascarulla@gmail.com}
  \icmlkeywords{causal discovery, natural language processing}

  \vskip 0.3in
]

\printAffiliationsAndNotice{}

\begin{abstract} 
  Discovering causal relations is fundamental to reasoning and intelligence.
  In particular, observational causal discovery algorithms estimate the
  cause-effect relation between two \emph{random entities} $X$ and $Y$, given
  $n$ samples from $P(X,Y)$. 
  
  In this paper, we develop a framework to estimate the cause-effect relation
  between two \emph{static entities} $x$ and $y$: for instance, 
  an art masterpiece $x$ and its fraudulent copy $y$. To this end, we
  introduce the notion of \emph{proxy variables}, which allow the construction of
  a pair of \emph{random entities} $(A,B)$ from the pair of static entities
  $(x,y)$. Then, estimating the cause-effect relation between $A$ and $B$ using
  an observational causal discovery algorithm leads to an estimation of the
  cause-effect relation between $x$ and $y$.  For example, our framework
  detects the causal relation between unprocessed photographs and their
  modifications, and orders in time a set of shuffled frames from a video.

  As our main case study, we introduce a human-elicited dataset of 10,000 pairs
  of casually-linked pairs of words from natural language.  Our methods 
  discover 75\percent{} of these causal relations. Finally, we discuss
  the role of proxy variables in machine
  learning, as a general tool to incorporate static knowledge 
  into prediction tasks.
\end{abstract} 

\section{Introduction}\label{sec:intro}

Discovering causal relations is a central task in science
\citep{pearl_causality:_2009, beebee2009oxford},  
and empowers humans to explain their experiences, predict the outcome of their
interventions, wonder about what could have happened but never did, or plan
which decisions will shape the future to their maximum benefit. Causal discovery is 
essential to the development of common-sense
\citep{kuipers1984commonsense,waldrop1987causality}.  In machine learning, it
has been argued that causal discovery algorithms are a necessary step towards
machine reasoning \citep{bottou2014machine, bottou2013counterfactual, dlp-phd}
and artificial intelligence \citep{lake2016building}. 

The gold standard to discover causal relations is to perform active
interventions (also called experiments) in the system under study
\citep{pearl_causality:_2009}.  However, interventions are in many situations
expensive, unethical, or impossible to realize.  In all of these situations,
there is a prime need to discover and reason about causality purely from
observation.  Over the last decade, the state-of-the-art in \emph{observational
causal discovery} has matured into a wide array of algorithms
\citep{shimizu2006linear, hoyer2009nonlinear, daniusis2012inferring,
Peters2014anm, mooij2016distinguishing, rcc, dlp-phd}. All these algorithms
estimate the causal relations between the random variables $(X_1,\ldots,
X_p)$ by estimating various asymmetries in
$P(X_1,\ldots,X_p)$. In the interest of simplicity, 
this paper considers the problem of discovering the causal relation between two
variables $X$ and $Y$, given $n$ samples from $P(X,Y)$.

The methods mentioned estimate the causal relation between two \emph{random entities} $X$ and $Y$, but often we are interested instead in two
\emph{static entities} $x$ and $y$. These are a pair of single objects for which it is not possible to define a probability distribution directly.
Examples of such static entities may
include one art masterpiece and its fraudulent copy, one translated document and
its original version, or one pair of causally linked words in natural language,
such as ``virus'' and ``death''.
Looking into the distant future, an algorithm able to discover the causal structure
between static entities in natural language could read throughout
medical journals, and discover the causal mechanisms behind a new cure for a
specific disease--the very goal of the ongoing \$45 million dollar \emph{Big
Mechanism} DARPA initiative \citep{cohen2015darpa}. Or, if we were able to
establish the causal relation between two arbitrary natural language
statements, we could tackle general-AI tasks such as the Winograd schema
challenge \citep{levesque2012winograd}, which are out-of-reach for current
algorithms.  The above and many more are situations where causal
discovery between static entities is at demand.

\subsection*{Our Contributions}
First, we introduce the framework of \emph{proxy variables} to estimate the
causal relation between \emph{static entities}
(Section~\ref{sec:framework}). 

Second, we apply our framework to the task of inferring the cause-effect
relation between pairs of images (Section~\ref{sec:images}). In particular, our
methods are able to infer the causal relation between an image and its
stylistic modification in $80\percent{}$ of the cases,
and it can recover the correct ordering of a set of shuffled video frames
(Section~\ref{sec:images:exps}).

Third, we apply our framework to discover the cause-effect
relation between pairs of words in natural language
(Section~\ref{sec:language}).
To this end, we introduce a novel dataset of 10,000 human-elicited pairs of
words with known causal relation (Section~\ref{sec:language:dataset}). Our
methods are able to recover $75\percent{}$ of the cause-effect relations (such
as ``accident $\to$ injury'' or ``sentence $\to$ trial'') in this
challenging task (Section~\ref{sec:language:exps}).

Fourth, we discuss the role of \emph{proxy variables} as a tool to incorporate
external knowledge, as provided by static entities, into general prediction
problems (Section~\ref{sec:conclu}).

All our code and data are available at \texttt{anonymous}.

We start the exposition by introducing the basic language of observational
causal discovery, as well as motivating its role in machine
learning. 

\section{Causal Discovery in Machine Learning}\label{sec:background}

The goal of \emph{observational causal discovery} is to reveal the cause-effect
relation between two random variables $X$ and $Y$, given $n$ samples $(x_1,
y_1), \ldots (x_n, y_n)$ from $P(X,Y)$.  In particular, we say that ``$X$
causes $Y$'' if there exists a \emph{mechanism} $F$ that transforms the values
taken by the cause $X$ into the values taken by the effect $Y$, up to the
effects of some random noise $N$. Mathematically, we
write $Y \leftarrow F(X,N)$.  Such equation highlights an asymmetric assignment
rather than a symmetric equality. If we were to intervene and change
the value of the cause $X$, then a change in the value of the effect $Y$ would follow.
On the contrary, if we were to manipulate the value of the effect $Y$, a
change in the cause $X$ would not follow.

When two random variables share a causal relation, they often become
statistically dependent. However, when two random variables are statistically
dependent, they do not necessarily share a causal relation. This is at the
origin of the famous warning ``dependence does not imply causality''.
This relation between dependence and
causality was formalized by \citet{Reichenbach56} into the following principle.

\begin{principle}[Principle of common cause] 
\label{principle:cc}
If two random variables $X$ and $Y$ are statistically dependent ($X \not\perp
Y$), then one of the following causal explanations must hold: 
\begin{compactitem}
  \item[{i})] $X$ causes $Y$ (write $X \to Y$), or
  \item[{ii})] $Y$ causes $X$ (write $X \leftarrow Y$), or
  \item[{iii})] there exists a random variable $Z$ that is the common cause of
  both $X$ and $Y$ (write $X \leftarrow Z \to Y$). 
\end{compactitem}
In the third case, $X$ and $Y$ are conditionally independent given $Z$ (write
$X \independent Y \given Z$).
\end{principle}

\begin{figure}[t!]
  \begin{center}
  \begin{subfigure}{.49\linewidth}
    \centering
    \includegraphics[width=\textwidth]{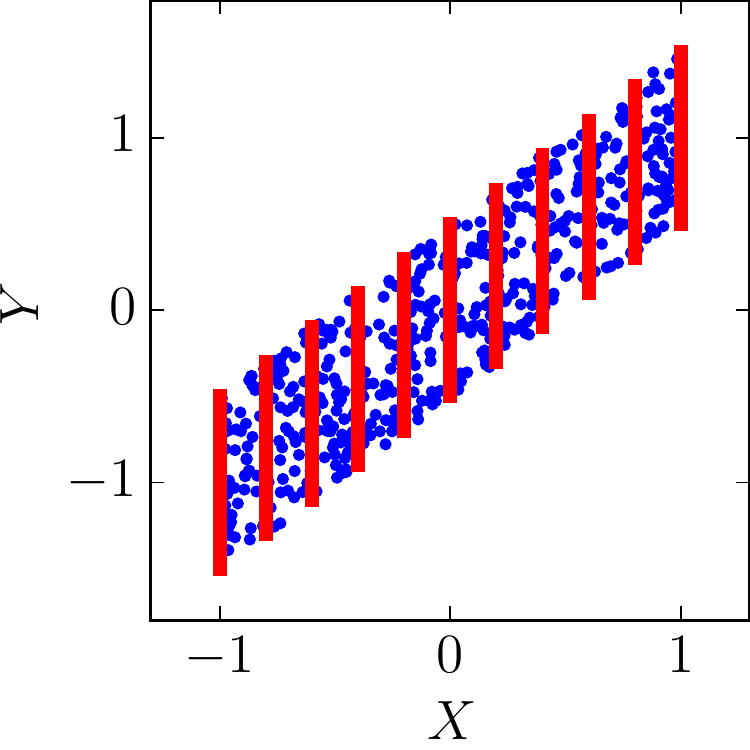}
    \caption{$Y = F(X)+N$, $X \perp N$.}
    \label{fig:footprints:anm:a}
  \end{subfigure}
  \hfill
  \begin{subfigure}{.49\linewidth}
    \centering
    \includegraphics[width=\textwidth]{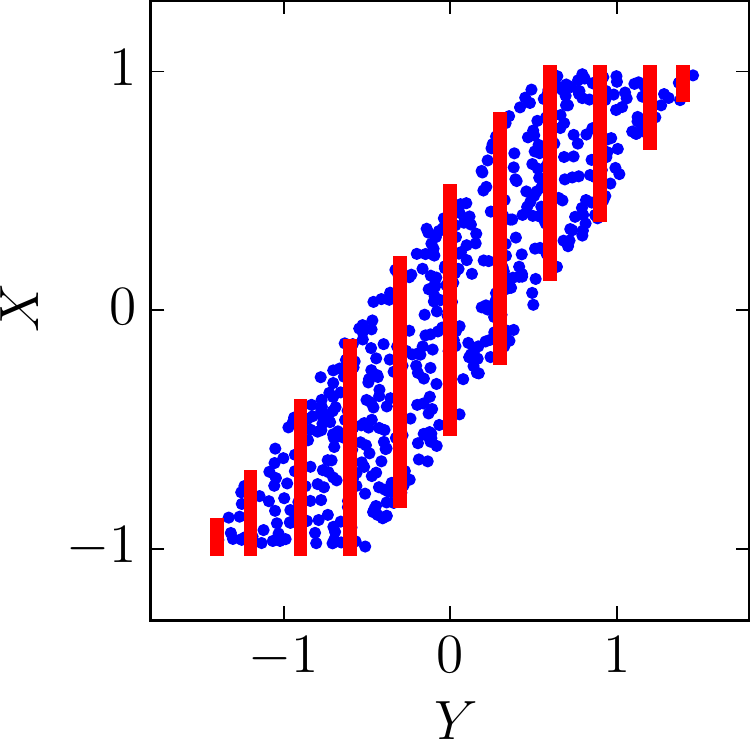}
    \caption{$X = G(Y)+E$, $Y \not\perp E$.}
    \label{fig:footprints:anm:b}
  \end{subfigure}
  \end{center}
  \caption{Example of an Additive Noise Model (ANM).}
  \label{fig:anm}
\end{figure}

In machine learning, these three types of statistical dependencies
are exploited without distinction, as dependence is sufficient to perform
optimal predictions about identically and independently distributed (iid) data
\citep{scholkopf2012causal}.  However, we argue that taking into account
the Principle of common cause would have far-reaching benefits in
non-iid machine learning. For example, assume that we are interested in
predicting the values of a target variable $Y$, given the values taken by
two features $(X_1, X_2)$. Then, understanding the causal structure underlying
$(X_1, X_2, Y)$ brings two benefits.
  
First, \emph{interpretability}.  Explanatory questions such as \emph{``Why does
$Y = 2$ when $(X_1, X_2) = (-1,3)$?''}, and counterfactual questions such as
\emph{``What value would have $Y$ taken, had $X_2 = -3$?''} cannot be answered
using statistics alone, since their answers depend on the particular causal
structure underlying the data.

Second, \emph{robustness}. Predictors which estimate the values taken by a
target variable $Y$ given only its direct causes are robust 
with respect to distributional shifts on their inputs.
For example, let $X_1 \sim {P}(X_1)$, $Y \leftarrow F_1(X_1)$, and $X_2
\leftarrow F_2(X_1)$. Then, the predictor $\BB{E}(Y \given X_1)$ is invariant
to changes in the joint distribution $P(X_1,X_2)$ as long as the causal
mechanism $F_1$ does not change.  However, the predictor $\BB{E}(Y \given X_1,
X_2)$ can vary wildly even if the causal mechanism $F_1$ (the only one involved
in computing $Y$) does not change \citep{peters_causal_2015,rojas2015}.

The previous two points apply to the common ``non-iid'' situations where we
have access to data drawn from some distribution ${P}$, but we are interested
in  some different but related distribution $\tilde{P}$. One natural way 
to phrase and leverage the similarities between $P$ and $\tilde{P}$ is in terms of
shared causal structures \citep{Peters2012,dlp-phd}.

While it is indeed an attractive endeavor, discovering the causal relation
between two random variables purely from observation is an impossible task 
when considered in full
generality. Indeed, any of the three causal structures outlined in 
Principle~\ref{principle:cc} could explain the observed dependency between 
two random variables. 
However,
one can in many cases impose assumptions to render the causal relation between
two variables \emph{identifiable} from their joint distribution.  For example,
consider the family of Additive Noise Models, or ANM \citep{hoyer2009nonlinear,
Peters2014anm, mooij2016distinguishing}. In ANM, one assumes that the causal
model has the form $Y= F(X) + N$, where $X \perp N$. It turns out that, under
some assumptions, the reverse ANM $X = G(Y) + E$ will not satisfy the
independence assumption $Y \perp E$ (Fig.~\ref{fig:anm}). The statistical
dependence shared by the cause and noise in the wrong causal direction is the
\emph{footprint} that renders the causal relation between $X$ and $Y$
\emph{identifiable} from statistics alone.

In situations where the ANM assumption is not satisfied (e.g., multiplicative
or heteroskedastic noise) one may prefer learning-based causal discovery tools,
such as the Randomized Causation Coefficient \citep{rcc}. RCC assumes access to
a causal dataset $D = \{ (S_i, l_i)\}_{i=1}^n$, where $S_i = (x_{i,j},
y_{i,j})_{j=1}^{n_i} \sim P^i(X_i, Y_i)$ is a bag of examples drawn from some
distribution $P^i$, $\ell_i = +1$ if $X_i \to Y_i$, and $\ell_i = -1$ if $X_i
\leftarrow Y_i$. By featurizing each of the training distribution samples $S_i$
using kernel mean embeddings \citep{smola2007hilbert}, RCC learns a binary
classifier on $D$ to reveal the causal footprints necessary to classify new
pairs of random variables.

However, both ANM and RCC based methods need $n \gg 1$ samples from $P(X,Y)$ to
classify the causal relation between the random variables $X$ and $Y$.
Therefore, these methods are not suited to infer the causal relation between
\emph{static entities} such as, for instance, one painting and its fraudulent
copy. In the following section, we propose a framework to extend the
state-of-the-art in causal discovery methods to this important case.

\section{The Main Concepts: Static Entities, Proxy Variables and Proxy
Projections}\label{sec:framework}

In the following, we consider two \emph{static entities} $x,y$ in some space $\mathcal{S}$
that satisfy the relation ``$x$ causes $y$''. 
Formally, this causal relation manifests the existence of a (possibly noisy) mechanism $f$ 
such that the value $y$ is computed as 
$y\leftarrow f(x)$. This asymmetric assignment guarantees changes in the \emph{static cause}
$x$ would lead to changes in the \emph{static effect} $y$, but the converse would not hold.

As mentioned previously, traditional causal discovery methods cannot be directly applied to 
static entities. In order to discover the causal relation between the pair of static entities $x$ and
$y$, we introduce two main concepts: \emph{proxy variables} $W$, and
\emph{proxy projections} $\pi$.

First, a \emph{proxy random variable} $W$ is a random variable taking values in
some set $\mathcal{W}$, which can be understood as a random source of information related to $x$ and $y$. 
This definition is  on purpose rather vague and will be illustrated through several examples in the following sections.

Second, a \emph{proxy projection} is a function $\pi : \mathcal{W} \times
\mathcal{S} \to \mathbb{R}$. Using a proxy variable and projection, we can
construct a pair of scalar random variables $A = \pi(W,x)$ and $B = \pi(W,y)$. 
A proxy variable and projection are \emph{causal} if the pair of random
entities $(A,B)$ share the same \emph{causal footprint} as the pair of static
entities $(x,y)$.\footnote{The concept of causal footprint is relative to our
assumptions. For instance, when assuming an ANM $Y \leftarrow f(X)+N$, the
causal footprint is the statistical independence between $X$ and $N$.} 

\begin{figure}
\begin{center}
\begin{tikzpicture}
  \draw (0,3) node(x2)  [vertex, rectangle, minimum width=2em] {$x$};
  \draw (4,3) node(x3)  [vertex, rectangle, minimum width=2em] {$y$};
  \draw (2,2) node(c)   [vertex, ] {$W$};
  \draw (0,1) node(a) [vertex, ] {$A$};
  \draw (4,1) node(b) [vertex, ] {$B$};
  \draw[edge, blue, very thick] (x2) -- (x3);
  \draw[edge, blue, dotted, very thick] (a) -- (b);
  \draw[edge] (x2) -- (a);
  \draw[edge] (x3) -- (b);
  \draw[edge] (c) -- (a);
  \draw[edge] (c) -- (b);
  \draw (3.75,1.75) node(b) [] {$\pi$};
  \draw (0.25,1.75) node(b) [] {$\pi$};
\end{tikzpicture}
\caption{A pair of \emph{static entities} $(x,y)$ share a causal relation of
interest (thick blue arrow).  A \emph{proxy variable} $W$,
together with a \emph{proxy projection} $\pi$ produces the \emph{random
entities} $(A,B)$, that share the causal footprint of $(x,y)$, denoted by
the dotted blue arrow.}
\label{fig:proxies}
\end{center}
\end{figure}
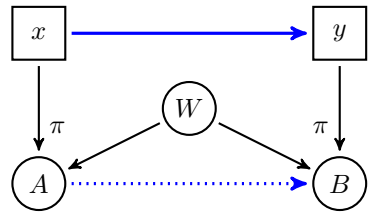

If the proxy variable and projection are causal, we may estimate the
cause-effect relation between the static entities $x$ and $y$ in three steps.
First, draw $(a_1, b_1), \ldots, (a_n, b_n)$ from $P(A,B)$. Second, use an
observational causal discovery algorithm to estimate the cause-effect relation
between $A$ and $B$ given $\{(a_i, b_i)\}_{i=1}^n$. Third, conclude ``$x$ causes
$y$'' if $A \to B$, or ``$y$ causes $x$'' if $A \leftarrow B$. This process is
summarized in Figure~\ref{fig:proxies}.

Note that the causal relation $X \to Y$ does not imply the causal relation $A
\to B$ in the \emph{interventional sense}: even if $A$ is a copy of $X$ and $B$
is a copy of $Y$, intervening on $A$ will not change $B$! We only care here about the presence of statistically observable
\emph{causal footprints} between the variables. Furthermore, our framework extends
readily to the case where $x$ and $y$ live in different modalities (say, $x$ is
an image and $y$ is a piece of audio describing the image). In this case, all we need is a proxy
variable $W = (W_x, W_y)$ and a pair of proxy projections $(\pi_x, \pi_y)$ with
the appropriate structure. 
For simplicity and throughout this paper, we will choose our proxy variables
and projections based on domain knowledge. Learning proxy variables and
projections from data is an exciting area left for future research.
\section{Causal Discovery Using Proxies in Images}\label{sec:images}

\begin{figure}[t!]
  \begin{center}
  \includegraphics[width=0.9\linewidth]{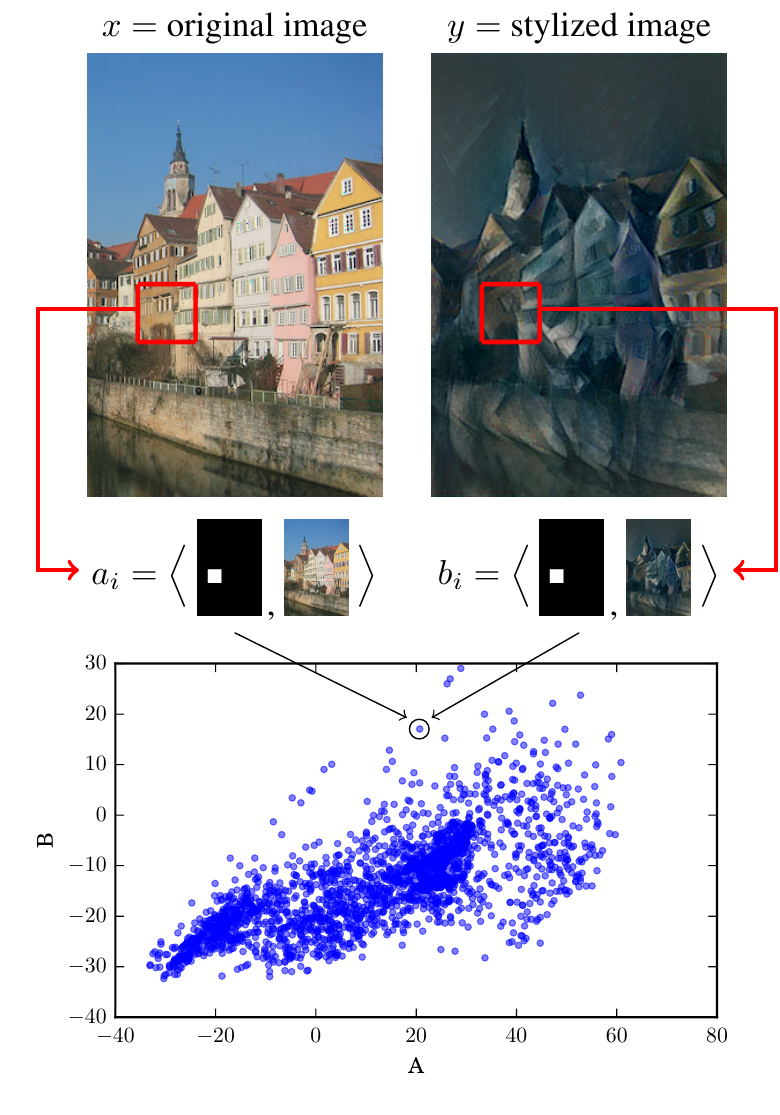}
  \end{center}
  \vspace*{-0.6cm}
  \caption{Sampling random patches at paired locations produces a proxy variable to
  discover the causal relation between two images.}
  \label{fig:tuebingen}
\end{figure}

Consider the two images shown in Figure~\ref{fig:tuebingen}.  The image on the
left is an unprocessed photograph of the T\"ubingen Neckarfront, while the one
on the right is the same photograph after being stylized with the algorithm of
\citet{gatys2016image}. From a causal point of view, the unprocessed image $x$ is the
cause of the stylized image $y$.  How can we leverage the ideas from 
Section~\ref{sec:framework} to recover such causal relation?

The following is one possible solution.  Assume that the two images are represented
by pixel intensity vectors $x$ and $y$, respectively. For $n \gg 1$ and 
$j = 1, \ldots,n$:
\begin{compactitem}
  \item Draw a mask-image $w_j$, which contains ones inside a patch at random
  coordinates, and zeroes elsewhere.
  \item Compute $a_j = \langle w_j, x \rangle$, and $b_j = \langle w_j, y
  \rangle$.
\end{compactitem}

This process returns a sample $\{(a_j, b_j)\}_{j=1}^{n}$ drawn from $P(A,B)$,
the joint distribution of the two scalar random variables $(A, B)$. The
conversion from static entities $(x,y)$ to random variables $(A, B)$ is
obtained by virtue of i) the randomness generated by the \emph{proxy variable}
$W$, which in this particular case is incarnated as random masks and ii) a 
causal projection $\pi$, here a simple dot product.

At this point, if the causal footprint between the random entities $(A, B)$
resembles the causal footprint between $x$ and $y$,  we can apply a regular
causal discovery algorithm to $(A,B)$ to estimate the causal relation
between $x$ and $y$. 

\subsection{Towards a Theoretical Understanding}
The intuition behind causal discovery using proxy variables is that, although
we observe $(x,y)$ as static entities, these
are underlyingly complex, high-dimensional, structured objects that carry rich
information about their causal relation. The proxy
variable $W$ introduces randomness to sample different views of the
high-dimensional causal structures, and $\pi$ summarizes those views into 
scalar values. But why should the causal footprint of these summaries cue 
the causal relation between $x$ and $y$?

We formalize this question for the specific case of stylized images, where $x$ is the original 
image and $y$ its stylized version. Let the causal 
mechanism mapping $x$ to $y$ operate
\emph{locally}.  More precisely, assume that each $k$-subset 
$y_{S_i}$ in the stylized image is computed from the $k$-subset 
$x_{S_i}$ in the original image, as described by the ANM:
\begin{align*}
  y_{S_i} = f(x_{S_i}) + \epsilon_{S_i}.
\end{align*}
Then, the stylized image $y = F(x) + \epsilon$, where
$F(x)_{S_i} = f(x_{S_i})$.  For simplicity, assume that $f(x_S) = g(\beta x_S)$
where $\beta$ is a $k\times k$ matrix and $g$ acts element-wise. 
Then, let $P(W)$ be a distribution over masks extracting random $k$-subsets,
and let $\pi(\cdot, \cdot) = \langle \cdot, \cdot \rangle$, to obtain:
\begin{align*}
  A = \pi(W, x) &= \langle W, x \rangle,\\
  B = \pi(W, y) &= \langle W, y \rangle\\
              &= \sum_{j=1}^k g_j\left( \sum_{l=1}^k \beta_{jl}(x_S)_l \right) + N\\
              &= \sum_{j=1}^k g_j\left( \sum_{l=1}^k \alpha_j A \right) + N 
\end{align*}
where $N = \sum_{j=1}^k (\epsilon_S)_j$, and where we assume that $\beta$ is such 
that $\alpha_j = \beta_{jl}$ for all $j
\leq k$.  Since $A \perp N$, the pair $(A,B)$ also follows an ANM. We leave for future work the 
investigation on identifiability conditions for causal inference using proxy variables.

\subsection{Numerical Simulations} \label{sec:images:exps}
In order to illustrate the use of causal discovery using proxy variables in
images, we conducted two small experiments.  In these experiments, we extract
$n=1024$ square patches of size $k=10$ pixels, and use the Additive Noise Model
\citep{hoyer2009nonlinear} to estimate the causal relation between the
constructed scalar random variables $A$ and $B$.

First, we collected $14$ unprocessed images together with $34$ stylizations
(including the one from Figure~\ref{fig:tuebingen}), made using the algorithm
of~\citet{gatys2016image}. When applying causal discovery using proxy variables
to this dataset, we can correctly identify the correct direction of causation from the original image to its
stylized version in $80\percent{}$ of the cases. 

\begin{figure}[t!]
  \begin{center}
  \includegraphics[width=0.245\linewidth]{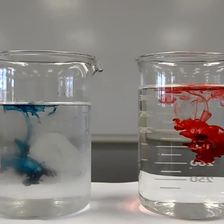}\hfill
  \includegraphics[width=0.245\linewidth]{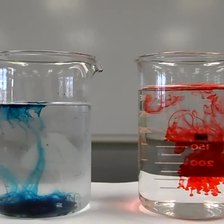}\hfill
  \includegraphics[width=0.245\linewidth]{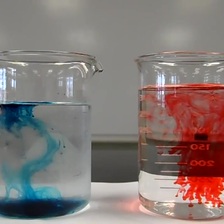}\hfill
  \includegraphics[width=0.245\linewidth]{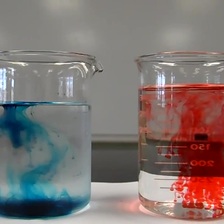}

  \includegraphics[width=0.245\linewidth]{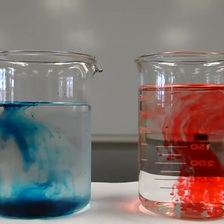}\hfill
  \includegraphics[width=0.245\linewidth]{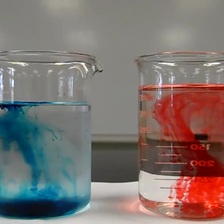}\hfill
  \includegraphics[width=0.245\linewidth]{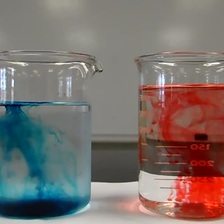}\hfill
  \includegraphics[width=0.245\linewidth]{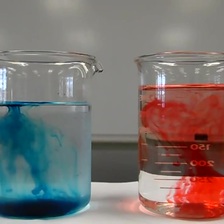}
  \end{center}
  \caption{Causal discovery using proxy variables uncovers the causal
  time signal to reorder a shuffled sequence of video frames.}
  \label{fig:frames}
\end{figure}

Second, we decomposed a video of drops of ink mixing with water into $8$
frames $\{(x_i)\}_{i=1}^8$, shown in Figure~\ref{fig:frames}. Using the same mask proxy variable as above, we construct an
$8\times 8$ matrix $M$ such that $M_{ij}=1$ if $x_i \rightarrow x_j$ according to our method and
$M_{ij} = 0$ otherwise. Then, we consider $M$ to be the adjacency matrix of the
causal DAG describing the causal structure between the $8$
frames. By employing topological sort on this graph, we were able to obtain the
true ordering, unique among the $40,320$ possible orderings.

\section{Causal Discovery Using Proxies in NLP}\label{sec:language}
As our main case study, consider discovering the
causal relation between pairs of words appearing in a large corpus of natural
language.  For instance, given the pair of words (virus, death), which represent 
our static entities $x$ and $y$, 
together with a large corpus of natural language, we want to recover causal
relations such as ``virus $\to$ death'', ``sun $\to$
radiation'', ``trial $\to$ sentence'', or ``drugs $\to$ addiction''.

This problem is extremely challenging for two reasons. First, word pairs are
extremely varied in nature (compare ``avocado causes guacamole'' to ``cat
causes purr''), and some are very rare (``wrestler causes pin'').
Second, the causal relation between two words can always be tweaked in
context-specific ways. For instance, one can construct
sentences where ``virus causes death'' (e.g., \emph{the virus led to
  a quick death}), but also sentences where ``death causes virus''
(e.g., \emph{the multiple deaths in the area further spread the
  virus}). We are hereby interested in the \emph{canonical} causal
relation between pairs of words, assumed by human subjects when
specific contexts are not provided (see
Section~\ref{sec:language:dataset}). Furthermore, our interest lies in
discovering the causal relation between pairs of words without the
use of language-specific knowledge or heuristics. To the contrary, we
aim to discover such causal relations by using generic observational
causal discovery methods, such as the ones described in
Section~\ref{sec:background}.

In the following, Section~\ref{sec:language:proxies} frames this problem in the
language of causal discovery between static entities. Then,
Section~\ref{sec:language:dataset} introduces a novel, human-generated,
human-validated dataset to test our methods. 
Section~\ref{sec:language:prior} reviews prior work on causal discovery in
language. Finally, Section~\ref{sec:language:exps} presents experiments evaluating our methods. 

\subsection{Static Entities, Proxies, and Projections for NLP}
\label{sec:language:proxies}

In the language of causal discovery with proxies, a pair of words is a pair of
static entities: $(x,y) = (\text{virus}, \text{death})$.
In order to discover the causal relation between $x$ and $y$, we are in need of
a proxy variable $W$, as introduced in Section~\ref{sec:framework}.  We will use a
simple proxy: let $P(W=w)$ be the probability of the word $w$ appearing in a
sentence drawn at random from a large corpus of natural language. 

Using the proxy $W$, we need to define the pair of random variables $A=
\pi(W,x)$ and $B=\pi(W,y)$ in terms of a causal projection $\pi$. Once we have
defined the causal projection $\pi$, we can sample $w_1, \ldots, w_n \sim
P(W)$, construct $a_i = \pi(w_i,x)$, $b_i = \pi(w_i,y)$, and apply a causal
discovery algorithm to the sample $\{(a_i, b_i)\}_{i=1}^n$. Specifically, we estimate $P(W)$ from a large corpus of natural language,
and sample $n=10,000$ words without replacement.\footnote{This is equivalent
to sampling approximately the top 10,000 most frequent words in the
corpus. Due to the extremely skewed nature of word frequency
distributions \citep{Baayen:2001}, sampling with replacement would produce 
a list of very frequent words such as \emph{a} and \emph{the}, sampled many
times.}

Throughout our experimental evaluation, we will use and compare different
proxy projections $\pi(w,x)$:
\begin{compactitem}
 \item[1)] $\pi_{\text{w2vii}}(w,x) = \dot{v^i_w}{v^i_x}$, where $v^i_z \in
 \mathbb{R}^d$ is the \emph{input} word2vec representation
 \citep{mikolov2013efficient} of the word $z$. The dot-product
 $\dot{v^i_w}{v^i_x}$ measures the \emph{similarity in meaning} between the
 pair of words $(w,x)$.
 \item[2)] $\pi_{\text{w2vio}}(w,x) = \dot{v^i_w}{v^o_x}$, where $v^o_z \in
 \mathbb{R}^d$ is the \emph{output} word2vec representation of the word $z$.
 The dot-product $\dot{v^i_w}{v^o_x}$ is an unnormalized estimate of the
 conditional probability $p(x|w)$ \citep{melamud2015simple}. 
 \item[3)] $\pi_{\text{w2voi}}(w,x) = \dot{v^o_w}{v^i_x}$, an unnormalized
 estimate of the conditional probability $p(w|x)$.
 \item[4)] $\pi_{\text{counts}}(w,x) = p(w,x)$, where the pmf $p(w,x)$ is
directly estimated from counting within-sentence co-occurrences in the corpus.
 \item[5)] $\pi_{\text{prec-counts}}(w,x)$ similar to the one above, but computed
 only over sentences where $w$ precedes $x$.
\item[6)] $\pi_{\text{pmi}}(w,x) =
  p(w,x)/(p(w)p(x))$, where the pmfs $p(w)$, $p(x)$, and
  $p(w,x)$ are estimated from counting words and (sentence-based)
  co-occurrences in the corpus. The log of this quantity is known as point-wise
  mutual information, or PMI \citep{church1990word}.
 \item[7)] $\pi_{\text{prec-pmi}}(w,x)$, similar to the one above, but computed
 only over sentences where $w$ precedes $x$.
\end{compactitem}

Applying the causal projections to our sample from proxy $W$, we construct the $n$-vector
\begin{align}
  \Pi_{\text{proj}}^x &= (\pi_{\text{proj}}(w_1, x), \ldots,
  \pi_{\text{proj}}(w_n, x))\label{eq:proxy_vector},
\end{align}
and similarly for $\Pi_{\text{proj}}^y$, where 
\begin{align}
  \text{proj} \in \{ &\text{w2vii}, \text{w2vio}, \text{w2voi},\nonumber\\
                     &\text{counts}, \text{prec-counts}, \text{pmi}, \text{prec-pmi} \}.\label{eq:allprojs}
\end{align}

In particular, we use the skip-gram model implementation of fastText
\citep{bojanowski2016ft} to compute $300-$dimensional word2vec representations.

\subsection{A Real-World Dataset of Cause-Effect Words}\label{sec:language:dataset}
We introduce a human-elicited, human-filtered dataset of $10,000$ pairs of
words with a known causal relation.  This dataset was constructed in two steps:
\begin{compactitem}
\item[1)] We asked workers from Amazon Mechanical Turk to create pairs of
words linked by a causal relation.  We provided the turks with examples of
words with a clear causal link (such as ``sun causes radiation'') and examples
of related words not sharing a causal relation (such as ``knife''
and ``fork''). For details, see Appendix~\ref{app:create}.
\item[2)] Each of the pairs collected from the previous step was randomly
shuffled and submitted to $20$ different turks, none
of whom had created any of the word pairs.  Each turk was required to classify
the pair of words $(x,y)$ as ``$x$ causes $y$'', ``$y$ causes $x$'', or ``$x$
and $y$ do not share a causal relation''. For more details, see
Appendix~\ref{app:validate}.
\end{compactitem}

This procedure resulted in a dataset of $10,000$ causal word
pairs $(x,y)$, each accompanied with three numbers: the number of turks that
voted ``$x$ causes $y$'', the number of turks that voted ``$y$ causes $x$'',
and the number of turks that voted ``$x$ and $y$ do not share a causal
relation''. 

\subsection{Causal Relation Discovery in NLP}\label{sec:language:prior}

The NLP community has devoted much attention to the problem of identifying the
semantic relation holding between two words, with causality as a special case.
\citet{Girju:etal:2009} discuss the results of the large shared task on
relation classification they organized (their benchmark included only 220 examples
of \emph{cause-effect}). The task required recognizing relations \emph{in
context}, but, as discussed by the authors, most contexts display the default
relation we are after here (e.g., ``The mutant \emph{virus} gave him a severe
\emph{flu}'' instantiates the default relation in which \emph{virus} is the
cause, \emph{flu} is the effect). All participating systems used extra
resources, such as ontologies and syntactic parsing, on top of corpus data.
They are thus outside the scope of the purely corpus-based methods we are
considering here. 

Most NLP work specifically focusing on the causality relation relies on
informative linking patterns co-occurring with the target pairs (such
as, most obviously, the conjunction \emph{because}). These patterns are
extracted and processed with sophisticated methods, involving annotation,
ontologies, bootstrapping and/or manual filtering (see, e.g.,
\citealt{Blanco:etal:2008} \citealt{Hashimoto:etal:2012},
\citealt{Radinsky:etal:2012}, and references therein). We experimented with
extracting linking patterns from our corpus, but, due
to the relatively small size of the latter, results were extremely sparse (note
that patterns can only be extracted from sentences in which both cause and
effect words occur). More recent work started looking at causal chains of
events as expressed in text (see \citealt{Mirza:Tonelli:2016} and references
therein). 
Applying our generic method to this task is a direction for future
work.

A semantic relation that received particular attention in NLP is that of
entailment between words (\emph{dog} entails \emph{animal}). As causality is
intuitively related to entailment, we will apply below entailment detection
methods to cause/effect classification. Most lexical entailment methods rely on
distributional representations of the words in the target pair. Traditionally,
entailing pairs have been identified with unsupervised asymmetric similarity
measures applied to distributed word representations
\citep{Geffet:Dagan:2005,Kotlerman:etal:2010,Lenci:Benotto:2012,Weeds:etal:2004}.
We will test  one of these related measures, namely, \emph{Weeds Precision
(WS)}. More recently, \citealt{Santus:etal:2014} showed that the relative
entropy of distributed vectors representing the words in a pair is an effective
cue to which word is entailing the other, and we also look at entropy for our
task. However, the most effective method to detect entailment is to apply a
supervised classifier to the concatenation of the vectors representing the
words in a pair \citep{Baroni:etal:2012,Roller:etal:2014,Weeds:etal:2014}.

\subsection{Experiments}\label{sec:language:exps}
We evaluate a variety of methods to discover the causal relation between two
words appearing in a large corpus of natural language. We study methods that fall within three categories:
\emph{baselines}, \emph{distribution-based causal discovery methods}, and
\emph{feature-based supervised methods}. These three families of methods
consider an increasing amount of information about the task at hand, and
therefore exhibit an increasing performance up to $85\percent{}$ classification
accuracy. 

All our computations will be based on the full English Wikipedia, as
post-processed by Matt Mahoney (see
{\small\url{http://www.mattmahoney.net/dc/textdata.html}}). We study the
$N=1,970$ pairs of words out of $10,000$ from the dataset described in
Section~\ref{sec:language:dataset} that achieved a consensus across at least
$18$ out of $20$ turks. We use RCC to estimate the
causal relation between pairs of random variables. 

\subsubsection{Baselines}\label{sec:baselines}
\newcommand\sxy{$S_{x\to y}$}
\newcommand\syx{$S_{x\leftarrow y}$}

These are a variety of unsupervised, heuristic baselines. Each baseline
computes two scores, denoted by \sxy{} and \syx{}, predicting $x \to y$
if $\text{\sxy{}} > \text{\syx{}}$, and $x \leftarrow y$ if
  $\text{\sxy{}} < \text{\syx{}}$. The baselines are:
\begin{compactitem}
  \item \emph{frequency}: \sxy{} is the number of sentences where $x$
  appears in the corpus, and \syx{} is the number of sentences
  where $y$ appears in the corpus.
  \item \emph{precedence}: considering only sentences from the corpus where
  both $x$ and $y$ appear, \sxy{} is the number of sentences where $x$
  occurs before $y$, and \syx is the number of sentences where
  $y$ occurs before $x$.
  \item \emph{counts (entropy)}: \sxy{} is the entropy of
  $\Pi_\text{counts}^x$, and \syx{} is the entropy of
  $\Pi_\text{counts}^y$, as defined in \eqref{eq:proxy_vector}.
  \item \emph{counts (WS)}: Using the WS measure of
  \citet{weeds2003general}, $\text{\sxy{}} = \text{WS}(\Pi_\text{counts}^x,
  \Pi_\text{counts}^y)$, and $\text{\syx{}} = \text{WS}(\Pi_\text{counts}^y,
  \Pi_\text{counts}^x)$.
  \item \emph{prec-counts (entropy)}: \sxy{} is the entropy of
  $\Pi_\text{prec-counts}^x$, and \syx{} is the entropy of
  $\Pi_\text{prec-counts}^y$ \eqref{eq:proxy_vector}.
  \item \emph{prec-counts (WS)}: analogous to the previous.
\end{compactitem}

The baselines \emph{PMI (entropy)}, \emph{PMI (WS)}, \emph{prec-PMI (entropy)}, \emph{prec-PMI (WS)} are analogous to the last
four, but use $(\Pi_\text{(prec-)pmi}^x, \Pi_\text{(prec-)pmi}^y)$ instead
of $(\Pi_\text{(prec-)counts}^x, \Pi_\text{(prec-)counts}^y)$. Figure~\ref{fig:results} shows
the performance of these baselines in blue.

\subsubsection{Distribution-based causal discovery methods}\label{sec:projections}
\textbf{These methods implement our framework of causal discovery using proxy variables.}
They classify $n$ samples from a 2-dimensional
\emph{probability distribution} as a whole.  Recall that a vocabulary 
$(w_j)_{j=1}^n$ drawn from the proxy is available.
Given $N$ word pairs $(x_i, y_i)$, this family of methods constructs a dataset
$D = \left\lbrace (\{(a_{j}^i,b_{j}^i)\}_{j=1}^n, \ell^i)\right\rbrace_{i=1}^N$, where $a_{j}^i =
\pi_\text{proj}({w}_j, x_i)$, $b_{j}^i = \pi_\text{proj}({w}_j, y_i)$,
$\ell^i = +1$ if $x_i \to y_i$ and $\ell^i = -1$ otherwise. In short, $D$ is a
dataset of $N$ ``scatterplots'' annotated with binary labels. The $i$-th
scatterplot contains $n$ 2-dimensional points, which are obtained by applying
the causal projection to both $x_i$ and $y_i$, against the $n$ vocabulary words
drawn from the proxy. 

The samples $(a_j^i, b_j^i)_{j=1}^n$ are computed using a deterministic
projection of iid draws from the proxy, meaning that $\{(a_j^i,
b_j^i)\}_{j=1}^n \sim P^n(A^i, B^i)$.  Therefore, we could permute the points
inside each scatterplot without altering the results of these methods.  In
principle, we could also remove some of the points in the scatterplot without a
significant drop in performance. Therefore, these methods search for causal
footprints \emph{at the 2-dimensional distribution level}, and we term them
\emph{distribution-based causal discovery methods}.

The methods in this family first split the dataset $D$ into a training set
$D_\text{tr}$ and a test set $D_\text{te}$. Then, the methods train 
RCC on the training set
$D_\text{tr}$, and test its classification accuracy on $D_\text{te}$. This
process is repeated ten times, splitting at random $D$ into a training set
containing $75\percent{}$ of the pairs, and a test set containing
$25\percent{}$ of the pairs. Each method builds on top of a causal projection
from \eqref{eq:allprojs} above.  Figure~\ref{fig:results} shows the test accuracy of
these methods in green.

\subsubsection{Feature-based supervised methods}\label{sec:supervised}

\begin{figure}
  \begin{center}
  \includegraphics[width=0.95\linewidth]{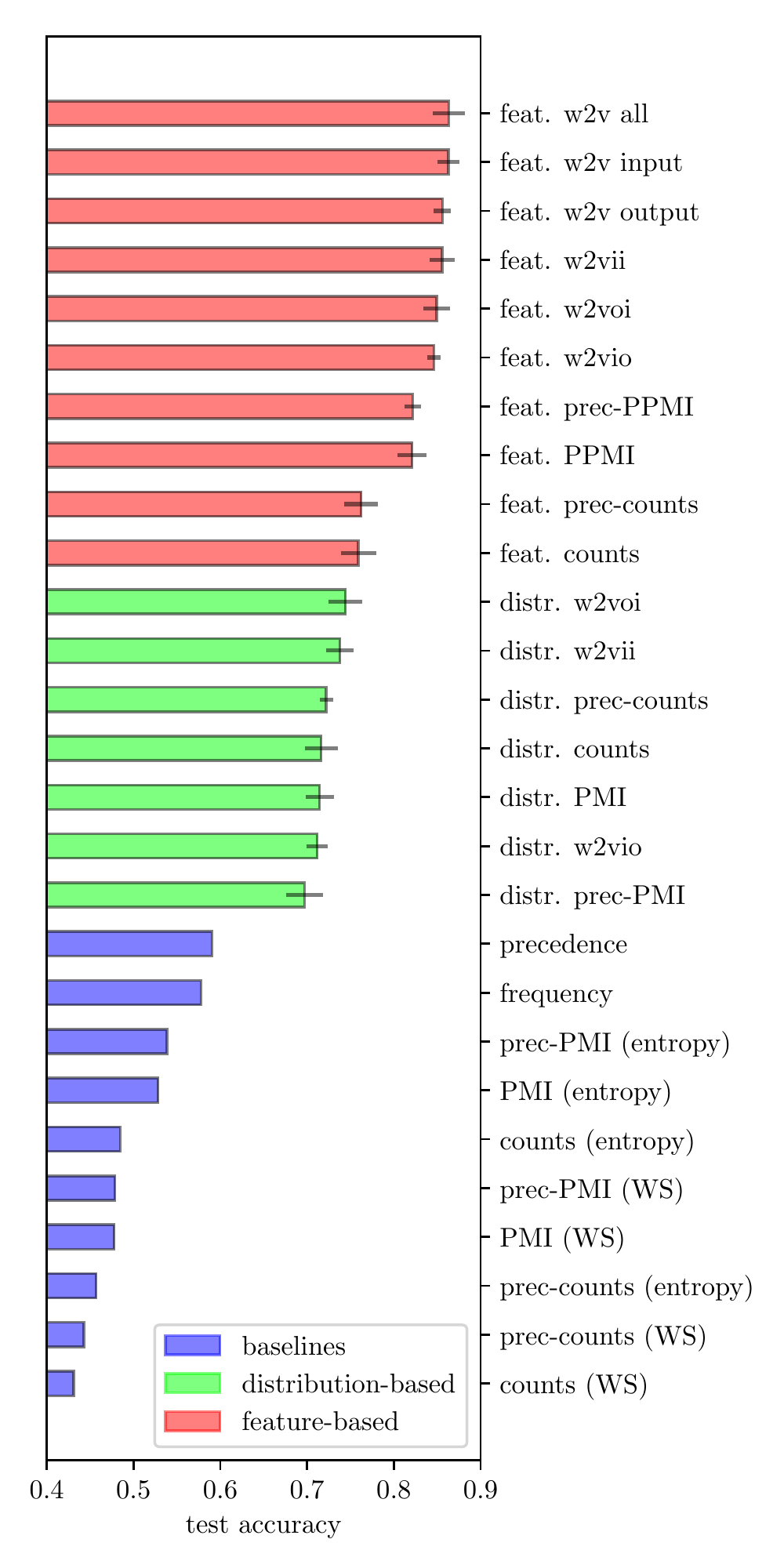}
  \end{center}
  \caption{Results for all methods on the NLP experiment. Accuracies above $52\percent{}$ are statistically significant
  with respect to a Binomial test at a significance level $\alpha = 0.05$.}
  \label{fig:results}
\end{figure}

These methods use the same data generated by our causal projections, but treat
them as fixed-size vectors fed to a generic classifier, rather than 
random samples to be analyzed with an observational causal discovery method. They
can be seen as an oracle to upper-bound the amount of causal signals (and signals correlated to causality) 
contained in our data. Specifically, they
use $2n$-dimensional vectors given by the concatenation of those
in~\eqref{eq:proxy_vector}.  Given $N$ word pairs $(x_i, y_i)$, they build a
dataset $D =  \left((\Pi_\text{proj}^{x_i}, \Pi_\text{proj}^{y_i}),
\ell^i\right)_{i=1}^N$, where $\ell^i = +1$ if $x_i \to y_i$, $\ell^i = -1$ if
$x_i \leftarrow y_i$, and ``proj'' is a projection from \eqref{eq:allprojs}.
Next, we split the dataset $D$ into a training set $D_\text{tr}$ containing
$75\percent{}$ of the pairs, and a disjoint test set $D_\text{te}$ containing
$25\percent{}$ of the pairs. To evaluate the accuracy of each method in this
family, we train a random forest of $500$ trees using $D_\text{tr}$, and report
its classification accuracy over $D_\text{te}$.  This process is repeated ten
times, by splitting the dataset $D$ at random.  The results are presented as
red bars in Figure~\ref{fig:results}. We also report the classification
accuracy of training the random forest on the raw word2vec
representations of the pair of words (top three bars).

\subsubsection{Discussion of results}

Baseline methods are the lowest performing, up to $59\percent{}$ test 
accuracy. We believe that the performance of the best baseline,
\emph{precedence}, is due to the fact that most Wikipedia is written in the active
voice, which often aligns with the temporal sequence of events, and thus
correlates with causality.

The feature-based methods perform best, achieving up to
$85\percent{}$ test classification accuracy. However, feature-based methods
enjoy the flexibility of considering each of the $n = 10,000$ elements in the
causal projection as a distinct feature.  Therefore, feature-based methods do
not focus on patterns to be found at a distributional level (such as
causality), and are vulnerable to permutation or removal of features.  We
believe that feature-based methods may achieve their superior performance by
overfitting to biases in our dataset, which are not necessarily
related to causality. 

\begin{figure}
  \begin{center}
  \includegraphics[width=0.95\linewidth]{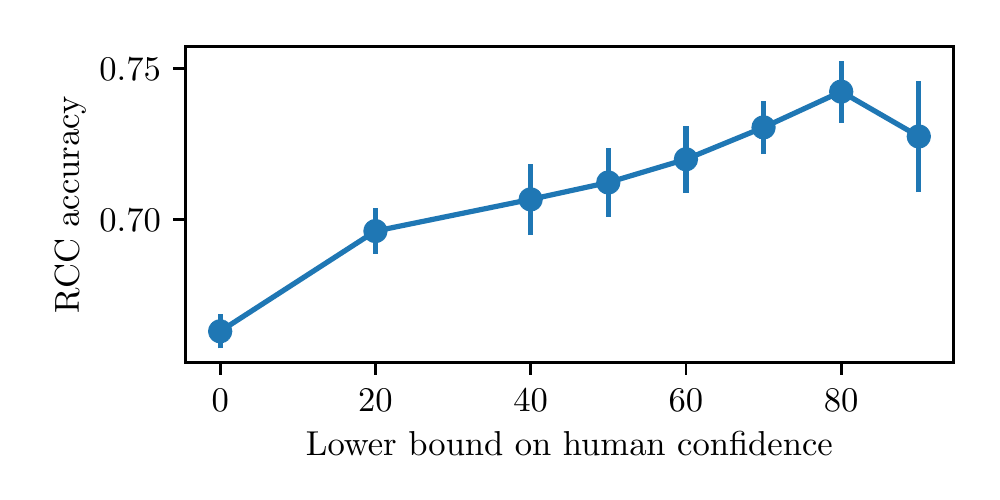}
  \end{center}
  \caption{RCC accuracy versus human confidence.}
  \label{fig:turks_rcc}
\end{figure}

Impressively, the best distribution-based causal discovery method achieves
$75\percent{}$ test classification accuracy, which is a significant improvement
over the best baseline method.  Importantly, our distribution-based methods
take a whole $2$-dimensional distribution as input to the classifier; as such,
these methods are robust with respect to permutations and removals of the $n$
distribution samples. We find it encouraging that the best distribution-based
method is the one based on $\pi_{\text{w2voi}}$. This suggests the intuitive
interpretation that the distribution of a vocabulary conditioned on
the cause word \emph{causes} the distribution of the vocabulary
conditioned on the effect word. Even more encouragingly,
Figure~\ref{fig:turks_rcc} shows a positive dependence between the test
classification accuracy of RCC and the confidence of human annotations, when
considering the test classification accuracy of all the causal pairs annotated
with a human confidence of at least $\{0, 20, 40, 50, 60, 70, 80, 90\}$. Thus,
our proxy variables and projections arguably capture a notion of causality aligned with the one of human annotators.

\section{Proxy Variables in Machine Learning}\label{sec:conclu}

The central concept in this paper is the one of \emph{proxy variable}.
This is a variable $W$ providing a random source of information related 
to $x$ and $y$. 

However, we can consider the reverse process of using a static entity $w$
to augment random statistics about a pair of random variables $X$ and
$Y$. As it turns out, this could be an useful process in general prediction problems.

To illustrate, consider a supervised learning problem mapping a
\emph{feature} random variable $X$ into a \emph{target} random variable $Y$.
Such problem is often solved by considering a sample $\{(x_i, y_i)\}_{i=1}^n
\sim P^n(X,Y)$. In this scenario, we may contemplate an unpaired, external,
\emph{static} source of information $w$ (such as a memory), which might help  solving
the supervised learning problem at hand. One could incorporate the information
in the static source $w$ by constructing the proxy projection $w_i =
\pi(x_i, w)$, and add them to the dataset to obtain $\{((x_i, w_i),
y_i)\}_{i=1}^n$ to build the predictor $f(x,\pi(x,w))$.

\section{Conclusion}

We have introduced the necessary machinery to estimate the causal relation
between pairs of static entities $x$ and $y$ --- one piece of art and its
forgery, one document and its translation, or the concepts underlying a
pair of words appearing in a corpus of natural language. We have done so
by introducing the tool of proxy variables and projections, reducing our problem
to one of observational causal inference between random entities.
Throughout a variety of experiments, we have shown the empirical
effectiveness of our proposed method, and we have connected it to the general
problem of incorporating external sources of knowledge as additional features in machine learning problems.

\clearpage
\newpage

\bibliographystyle{icml2017}
\bibliography{paper}

\clearpage
\newpage

\appendix
\onecolumn

\icmltitle{Supplementary material to \emph{Causal discovery using proxy variables}}

\section{Instructions for word pair creators}\label{app:create}

We will ask you to write word pairs (for instance, WordA and WordB) for which you believe the statement ``WordA causes WordB'' is true.\\

To provide us with high quality word pairs, we ask you to follow these indications:

\begin{itemize}
\item All word pairs must have the form ``WordA $\rightarrow$ WordB''. It is essential that the first word (WordA) is the cause, and the second word (WordB) is the effect.
\item WordA and WordB must be one word each (no spaces, and no ``recessive gene $\rightarrow$ red hair''). Avoid compound words such as ``snow-blind''.
\item In most situations, you may come up with a word pair that can be justified both as ``WordA $\rightarrow$ WordB'' and ``WordB $\rightarrow$ WordA''. In such situations, prefer the causal direction with the easiest explanation. For example, consider the word pair ``virus $\rightarrow$ death''. Most people would agree that ``virus causes death``. However, ``death causes virus'' can be true in some specific scenario (for example, ``because of all the deaths in the region, a new family of virus emerged.''). However, the explanation ``virus causes death`` is preferred, because it is more general and depends less on the context.
\item We do not accept word pairs with an ambiguous causal relation, such as ``book - paper''.
\item We do not accept simple variations of word pairs. For example, if you wrote down ``dog $\rightarrow$ bark'', we will not credit you for other pairs such as ``dogs $\rightarrow$ bark'' or ``dog $\rightarrow$ barking''. 
\item Use frequent words (avoid strange words such as ``clithridiate'').
\item Do not rely on our examples, and use your creativity. We are grateful if you come up with diverse word pairs!
Please do not add any numbers (for example, ``1 - dog $\rightarrow$ bark'').
For your guidance, we provide you examples of word pairs that belong to different categories. Please bear in mind that we will reward your creativity: therefore, focus on providing new word pairs with an evident causal direction, and do not limit yourself to the categories shown below.
\end{itemize}

\textbf{1) Physical phenomenon}: there exists a clear physical mechanism that explains why ``WordA $\rightarrow$ WordB''.

\begin{itemize}
\item sun $\rightarrow$ radiation (The sun is a source of radiation. If the sun were not present, then there would be no radiation.)
\item altitude $\rightarrow$ temperature
\item winter $\rightarrow$ cold
\item oil $\rightarrow$ energy 
\end{itemize}

\textbf{2) Events and consequences}: WordA is an action or event, and WordB is a consequence of that action or event.

\begin{itemize}
\item crime $\rightarrow$ punishment
\item accident $\rightarrow$ death
\item smoking $\rightarrow$ cancer
\item suicide $\rightarrow$ death
\item call $\rightarrow$ ring
\end{itemize}

\textbf{3) Creator and producer}: WordA is a creator or producer, WordB is the creation of the producer.

\begin{itemize}
\item writer $\rightarrow$ book (the creator is a person)
\item painter $\rightarrow$ painting
\item father $\rightarrow$ son
\item dog $\rightarrow$ bark
\item bacteria $\rightarrow$ sickness
\item pen $\rightarrow$ drawing (the creator is an object)
\item chef $\rightarrow$ food
\item instrument $\rightarrow$ music
\item bomb $\rightarrow$ destruction
\item virus $\rightarrow$ death
\end{itemize}

\textbf{4) Other categories! Up to you, please use your creativity!}

\begin{itemize}
\item fear $\rightarrow$ scream
\item age $\rightarrow$ salary 
\end{itemize}

\section{Instructions for word pair validators}\label{app:validate}

Please classify the relation between pairs of words A and B into one of three categories: either ``A causes B'', ``B causes A'', or ``Non-causal or unrelated''.

For example, given the pair of words ``virus and death'', the correct answer would be:

\begin{itemize}
\item virus causes death (correct);
\item death causes virus (wrong);
\item non-causal or unrelated (wrong).
\end{itemize}

Some of the pairs that will be presented are non-causal. This may happen if:

\begin{itemize}
\item The words are unrelated, like ``toilet and beach''.
\item The words are related, but there is no clear causal
  direction. This is the case of ``salad and lettuce'', since we can
  eat salad without lettuce, or eat lettuce in a burger.
\end{itemize}

To provide us with high quality categorization of word pairs, we ask you to follow these indications:

\begin{itemize}
\item Prefer the causal direction with the simplest explanation. Most people would agree that ``virus causes death''. However, ``death causes virus'' can be true in some specific scenario (for example, ``because of all the deaths in the region, a new virus emerged.''). However, the explanation ``virus causes death'' is preferred, because it is true in more general contexts.
\item If no direction is clearer, mark the pair as non-causal. Here, conservative is good!
\item Think twice before deciding. We will present the pairs in random order!
\end{itemize}

Please classify all the presented pairs. If one or more has not been answered, the whole batch will be invalid. \textbf{PLEASE DOUBLE CHECK THAT YOU HAVE ANSWERED ALL 40 WORD PAIRS.}

Examples of causal word pairs:

\begin{itemize}
\item  ``sun and radiation'': sun causes radiation
\item  ``energy and oil'': oil causes energy
\item  ``punishment and crime'': crime causes punishment
\item  ``instrument and music'': instrument causes music
\item  ``age and salary'': age causes salary
\end{itemize}

Examples of non-causal word pairs:

\begin{itemize}
\item ``video and games'': non-causal or unrelated
\item ``husband and wife'': non-causal or unrelated
\item ``salmon and shampoo'': non-causal or unrelated
\item ``knife and gun'': non-causal or unrelated
\item ``sport and soccer'': non-causal or unrelated
\end{itemize}

\end{document}